\documentclass[letterpaper, 10 pt, conference]{ieeeconf}

\IEEEoverridecommandlockouts

\usepackage[T1]{fontenc}
\usepackage{graphicx}
\usepackage{subcaption}
\usepackage{tikz}
\usepackage{url}
\usetikzlibrary{arrows,calc, patterns, positioning, shapes.geometric, decorations.pathreplacing,decorations.markings}

\title{\LARGE \bf
 Exploring Mechanically Self-Reconfiguring Robots\\ for Autonomous Design
}

\author{T{\o}nnes F. Nygaard$^{1}$, Charles P. Martin$^{1}$, Jim Torresen$^{1}$ and Kyrre Glette$^{1}$%
\thanks{*This work is partially supported by The Research Council
  of Norway as a part of the Engineering Predictability with
  Embodied Cognition (EPEC) project, under grant agreement
  240862.}
\thanks{$^{1}$Authors are with the Robotics and Intelligent Systems research group,
        University of Oslo, Norway,
        {\tt\small tonnesfn@ifi.uio.no}}%
}

\begin{document}

\maketitle
\thispagestyle{empty}
\pagestyle{empty}

\begin{abstract}

Evolutionary robotics has aimed to optimize robot control and morphology to produce better and more robust robots.
Most previous research only addresses optimization of control, and does this only in simulation.
We have developed a four-legged mammal-inspired robot that features a self-reconfiguring morphology.
In this paper, we discuss the possibilities opened up by being able to efficiently do experiments on a changing morphology in the real world.
We discuss present challenges for such a platform and potential experimental designs that could unlock new discoveries.
Finally, we place our robot in its context within general developments in the field of evolutionary robotics, and consider what advances the future might hold.

\end{abstract}


\section{Introduction}

Robots are used in more and more complex environments, and are expected to be able to adapt to changes and unknown situations.
The easiest and quickest way to adapt, is to change the control system of a robot, but for increased adaptation one should also change the body of the robot--its morphology--to better fit the environment or task at hand.

The theory of embodied cognition (EC) has recently gained traction in the robotics community, which states that control is not the only source of cognition, and in fact the body, environment, and interaction between these and the mind all contribute as cognitive resources \cite{wilson_frontiers13_embodied}.
Taking advantage of these concepts could lead to improved adaptivity, robustness, and versatility in robotic systems~\cite{Pfeifer2007}.

In contrast to the majority of work in the field of evolutionary robotics (ER), where simulations are used to automatically design robotic controllers or morphologies~\cite{Bongard2013},
Eiben argues for real-world experiments in his ``Grand Challenges for Evolutionary Robotics'' paper~\cite{eiben_j_FRONTIERS_grandchallenges}: Being able to switch from the current use of simulators as the primary evaluator with occasional hardware feedback, to evaluating robots in hardware with the occasional simulator feedback, will be a game changer comparable to how evolution transitioned from wetware to software when computers made it possible to simulate evolutionary processes for the first time.
Being able to now do it in hardware exposes the system to the same rich and noisy environment that natural evolution experienced, and he argues that only then will we be able to accurately investigate evolutionary processes or the interaction between the body, mind and environment through embodied cognition.

\begin{figure}[t] 
\vspace{2mm}
\centering
  \begin{subfigure}{0.23\textwidth}
    \includegraphics[width=1.0\textwidth]{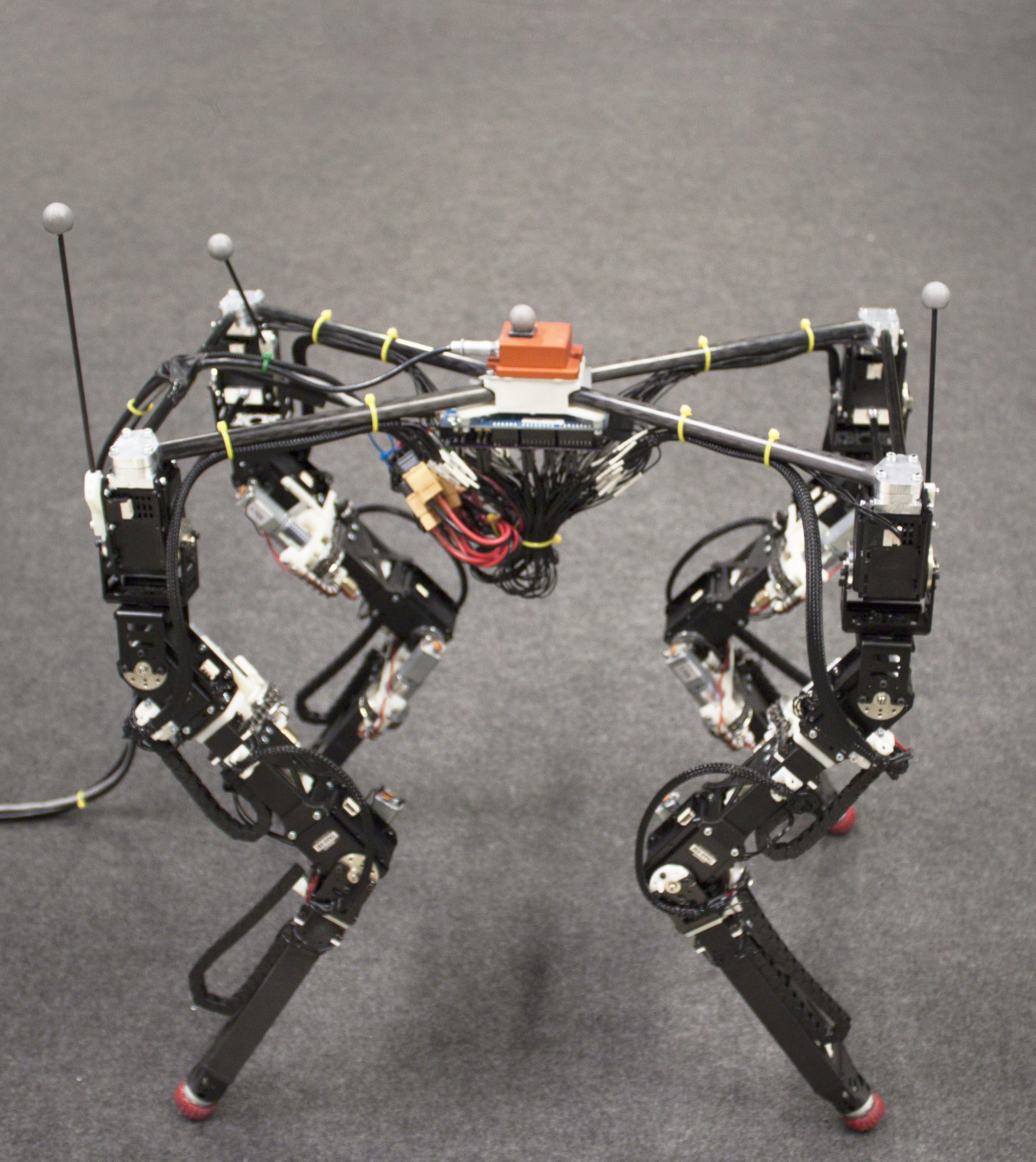}
  \end{subfigure}
  \begin{subfigure}{0.23\textwidth}
    \includegraphics[width=\textwidth]{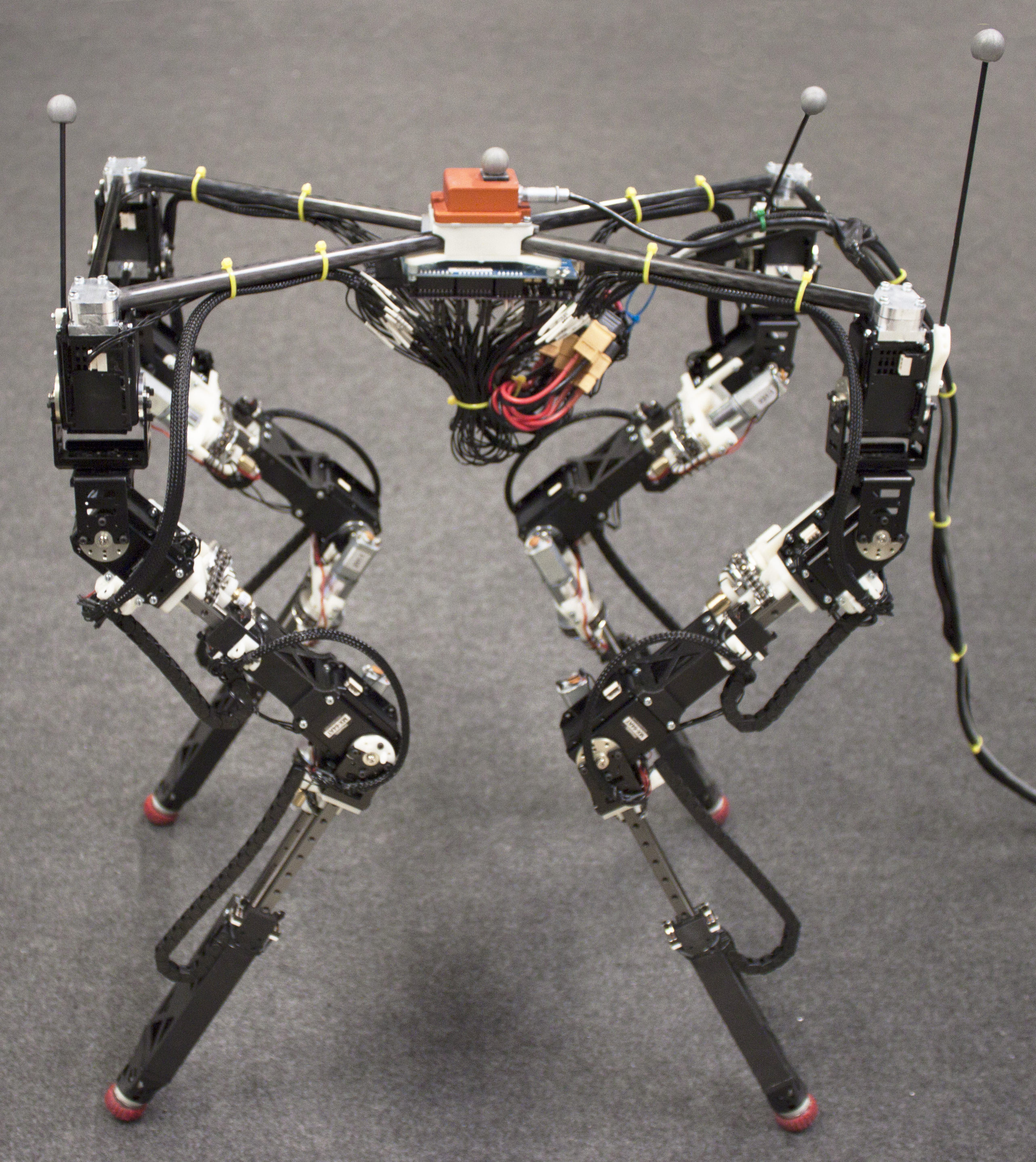}
  \end{subfigure}
  \caption{Our mechanically self-reconfiguring robot in its shortest (left) and longest possible leg configurations (right).}
  \label{fig.robot}
\vspace{-0.5cm}
\end{figure}

Along the same lines of thought, in this paper we discuss the concept of doing simultaneous optimization of control and morphology mainly through measurements on a real-world robot, in light of our mechanically self-reconfigurable robot platform (Fig.~\ref{fig.robot}) and currently performed experiments.
We also outline possible application areas of automatic design for robots with dynamic morphology.
We believe that the next step for better experimentation in the field of control-morphology co-evolution is to get a simple proof-of-concept example working for a self-reconfiguring robot, so that researchers can gather experience from real world experiments. 

\begin{figure*} 
  \centering
  \begin{subfigure}{0.58\textwidth}
    \includegraphics[width=\textwidth]{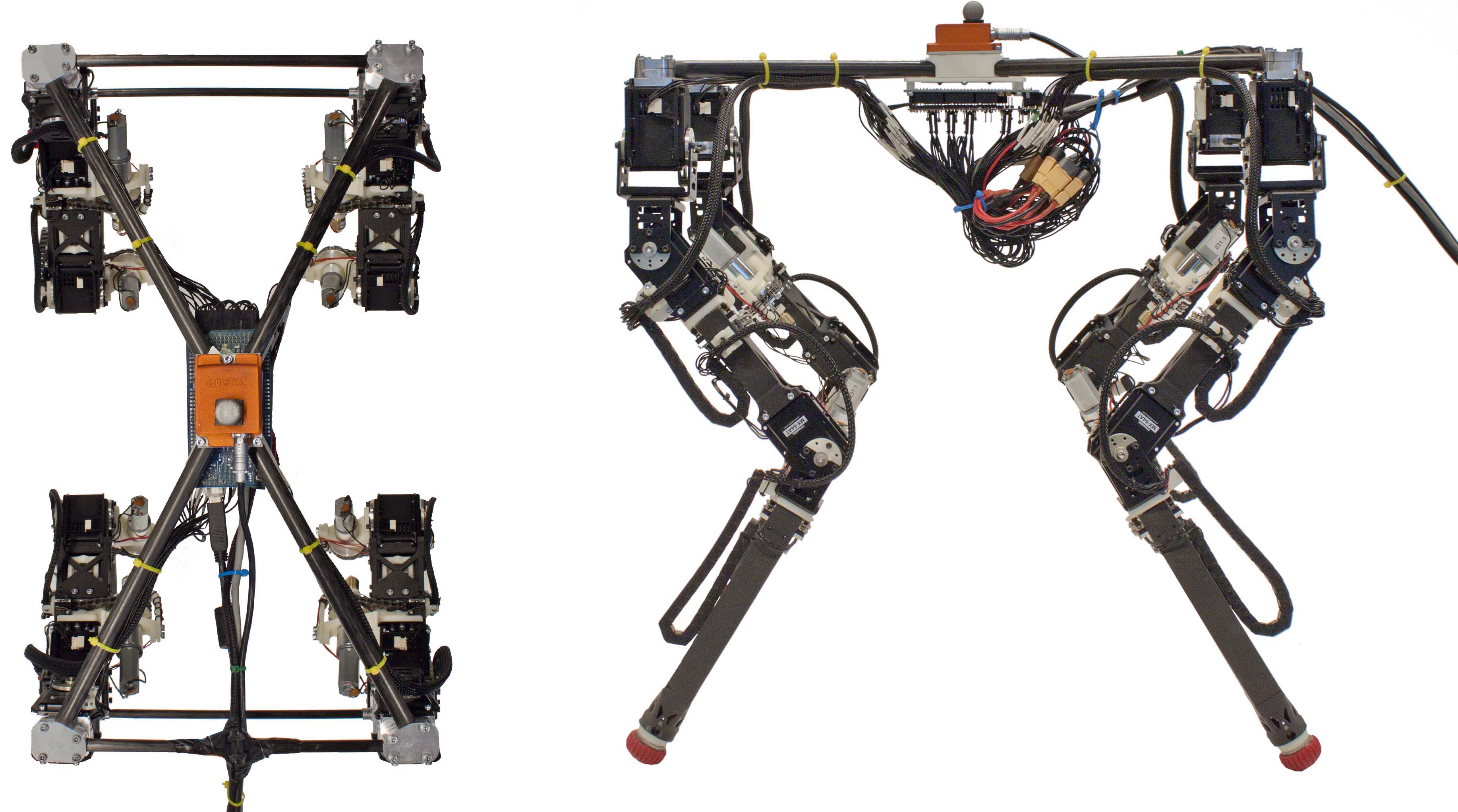}
  \end{subfigure}
  \hspace{5mm}
  \begin{subfigure}{0.24\textwidth}
    \includegraphics[width=\textwidth]{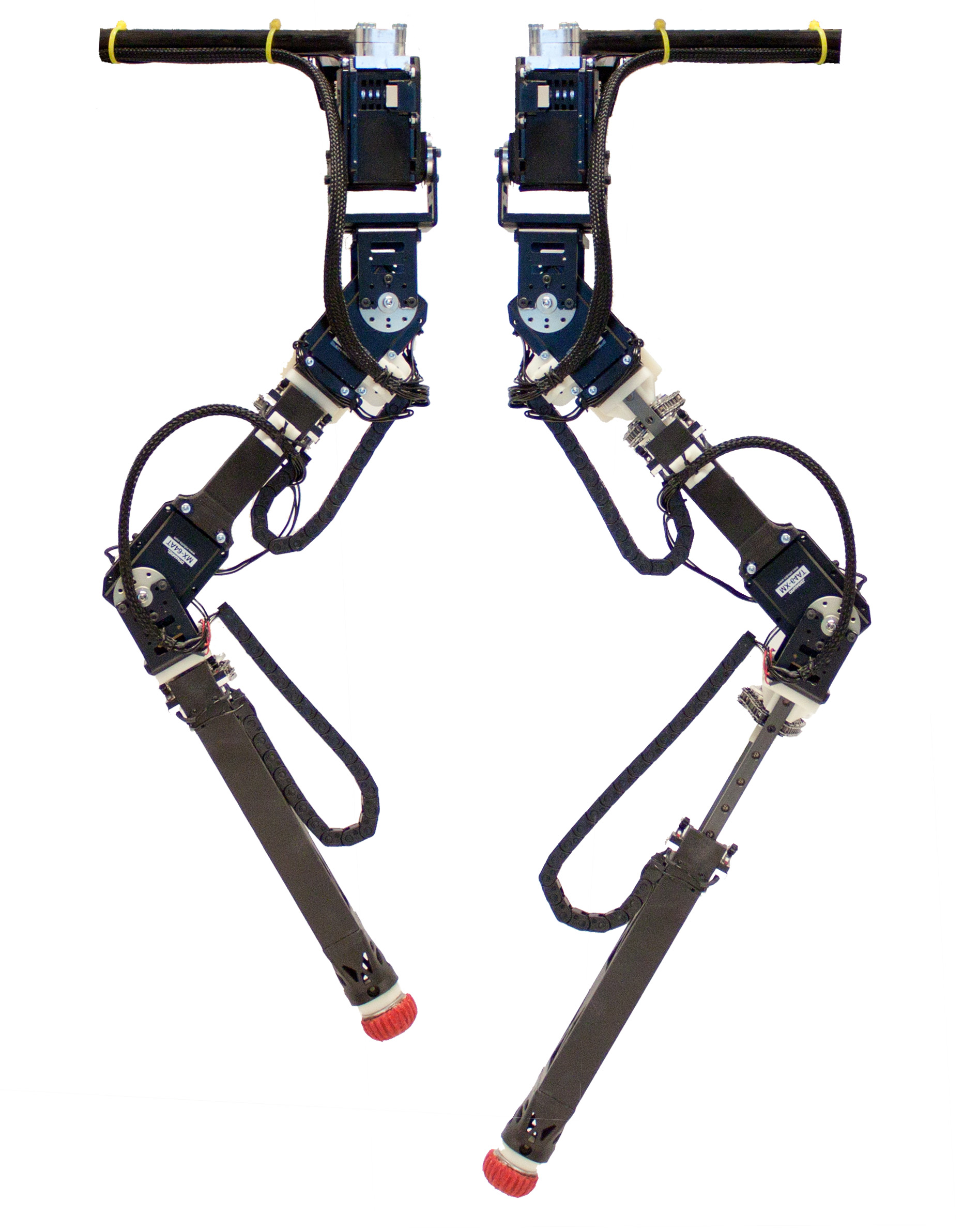}
  \end{subfigure}
  \caption{Top and left views of our reconfigurable robotic platform, and examples of the legs at two different lengths.}
  \label{fig.robotfig}
  \vspace{-0.3cm}
\end{figure*}


\section{Challenges and Applications}
A robot with reconfigurable morphology allows adaptation of both the body and the control, which gives higher flexibility compared to the usual case of a static morphology.
This will become increasingly important as robots are used in more complex environments, and work more in collaboration with other robots or humans.
Evolution of morphology and control would also allow for more experiments in the domain of embodied cognition.
In the following paragraphs we attempt to outline some challenges and possible applications when working with evolution for robots with reconfigurable morphology.

\subsection{Challenges}
One important aspect to consider is whether to base the evolutionary design process on evaluations in simulation, in hardware, or a mixture of these. 
Here we outline a few points to consider on this topic: 

\textbf{Reality gap.}
Most current work in evolution of morphology is carried out in simulation, with some transferring a select few morphologies to the real world by manually building them~\cite{lipson_j_NATURE00, tonnesfn_evorobot17}.
The effect known as the reality gap~\cite{jakobi_c_ALIFE95}, arising from discrepancies between the simulation and the real world, makes it very difficult to directly transfer solutions from evolution in simulation to real-world robots. 
The problem becomes even more difficult for robots with dynamic morphologies, making it necessary to tune the simulation for a range of morphologies.

\textbf{Environmental complexity.}
The study of embodied cognition relies on an accurate representation of the environment and the body, but also the interactions between these and the control system, which is an even more challenging task for physics simulation. 
Real world environments are richer and possess an inherent noise that is very hard, if not impossible, to replicate in simulations, and this could be especially important when studying embodied cognition effects.

\textbf{Search space complexity.}
While some previous work reports evolution of morphology in hardware alone, they require excessive human intervention~\cite{milan_c_ALIFE17}, or have very slow external reconfiguration~\cite{vujovic_c_ALIFE17}, which makes it difficult to evaluate enough solutions in the large and complex search space.
The increased size and complexity of the search space when evolving for morphology and control also makes the optimization more difficult, and the balance between exploration and exploitation is even more important to ensure early exploration without incurring premature convergence.
Cheney pointed out the difficulty of co-evolving morphology and control~\cite{cheney_ALIFE16}, arguing that mutations in the morphology effectively scrambles the interface between the controller and the environment.

\subsection{Applications} 
In this section we categorize some relevant applications for research using reconfigurable morphologies:

\textbf{Automatic design.} 
One promising application is to do automatic design which can take the robot morphology into account.
This is done using an optimization algorithm to change the body of the robot, and can be done in simulations or in the real world.
The result can be used directly, or an engineer can use the results of the optimizations and further improve it. 

\textbf{Environmental adaptation.} 
Another application that uses the body of a robot to adapt to new conditions is environmental adaptation -- by either adapting to internal conditions like changing actuator torque or a broken leg, or external conditions like a new walking surface. 

\textbf{Automatic evaluation.} 
A single physical robot is often used for evaluation, validation, and verification, but this can also be done for a range of robot bodies using a robot with dynamic morphology.
To generalize findings from design algorithms to a wider range of robots and applications, dynamic morphologies would make it easier to actually test on different robots than just testing one or two and interpolating the results to other robot bodies.
Even though evaluations in the real world take a lot of time, it can still be used efficiently with automatic evaluation with no requirements of human intervention, which opens up exploratory experiments on different robots.

\textbf{Meta-studies.} 
Another important application of dynamic morphology, is in the study of the search or optimization process itself, where we are not really looking for the best robot, but at the meta-statistics of the study.
This can be used to study evolution from a biological perspective, or high-level research into optimization techniques in realistic environments.

\textbf{Research goals.} 
It should be noted that different research fields have different end goals: 
Artificial life aims to replicate the natural processes of evolution by evolving populations of virtual creatures in realistic environments.
On the other hand, evolutionary robotics aims to automatically optimize robot bodies and their controllers, and allow the robots to adapt on-line to different unknown environments and tasks.


\section{Use of a self-modifying robot platform}
With the challenges of simulation-based approaches in mind, our work has initially been focused on a purely hardware-based approach.
Previous hardware-based solutions featuring reconfigurable morphologies have either required excessive human intervention to work, or slow external reconfiguration.
To surpass these, we decided to construct a mechanically self-reconfiguring robot, and decided that the length of the two bottom links in the legs would have the most impact as the reconfigurable elements. 
Pictures of the robot in its two extreme leg configurations can be seen in Fig.~\ref{fig.robot}, and details can be found in our previous work~\cite{tonnesfn_IROS18} or at our webpage\footnote{Please visit \url{http://robotikk.net/project/dyret/} for more information on the platform, including videos, source code, and design files.}.
To not increase the weight and inertia of the legs too much, quick and powerful motors were not feasible, and instead we selected powerful but slow motors to achieve leg length reconfiguration.
The slow motors mean that active use of the leg lengths as part of the locomotion would be too slow, rendering this in practice to a purely morphological reconfiguration mechanism.

\textbf{Technical details.}
Our platform was designed to be used for evolutionary experiments, and the robot is shown in Fig.~\ref{fig.robotfig}.
The whole robot weighs about 5.5kg, and is constructed of 3d-printed and off-the-shelf components.
It has three Dynamixel MX-64 servos in each leg for actuation, and two brushed DC motors for mechanical self-reconfiguration.
All software functions are implemented as separate Robot Operating System (ROS) nodes. 
Gaits are evaluated by walking both forwards and backwards, and each evaluation takes about one minute, including reconfiguration.
We use motion capture equipment to evaluate the distance walked, in addition to a stability measurement calculated from the on-board AHRS.

\subsection{Experiments so far with our platform}
\begin{figure}[t!]
  \centering
    \includegraphics[height=35mm]{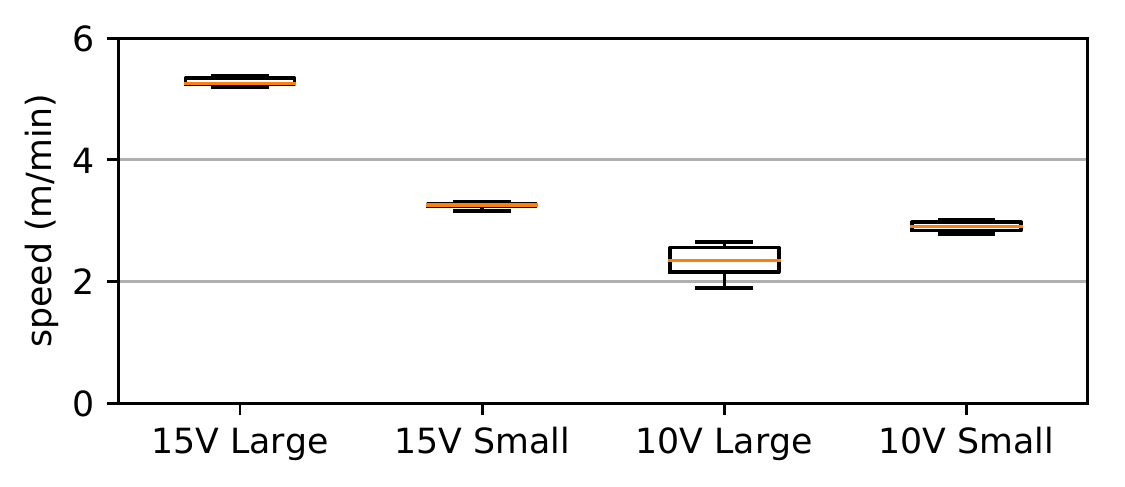}
    \caption{The performance of some hand-designed controller and morphology pairs for two different internal environments \cite{tonnesfn_IROS18}. We see the large robot performs best at 15V, while the smaller robot performs more consistently and better at 10V.}
    \label{fig.expresults1}
	\vspace{-0.5cm}
\end{figure}

\textbf{Validation experiments.}
When presenting our new reconfigurable platform~\cite{tonnesfn_IROS18}, we did preliminary experiments with hand designed gaits.
First, we chose two different morphologies and hand designed gaits to fit the two, with the largest robot being the fastest in optimal lab conditions.
Tests in the lab showed that by changing the internal environment of the robot, by reducing the available torque of the servos, the smaller morphology was suddenly the fastest, more optimal one.
This is shown in in Fig.~\ref{fig.expresults1}.
We also did some preliminary field studies with the robot, and observed that the large morphology performed best in a covered garage, but out in the Norwegian winter on icy conditions, the smaller robot did better.
By doing this, we showed that the robot achieves better performance in different environments by having a dynamic morphology.

\begin{figure}[t!]
    \includegraphics[height=35mm]{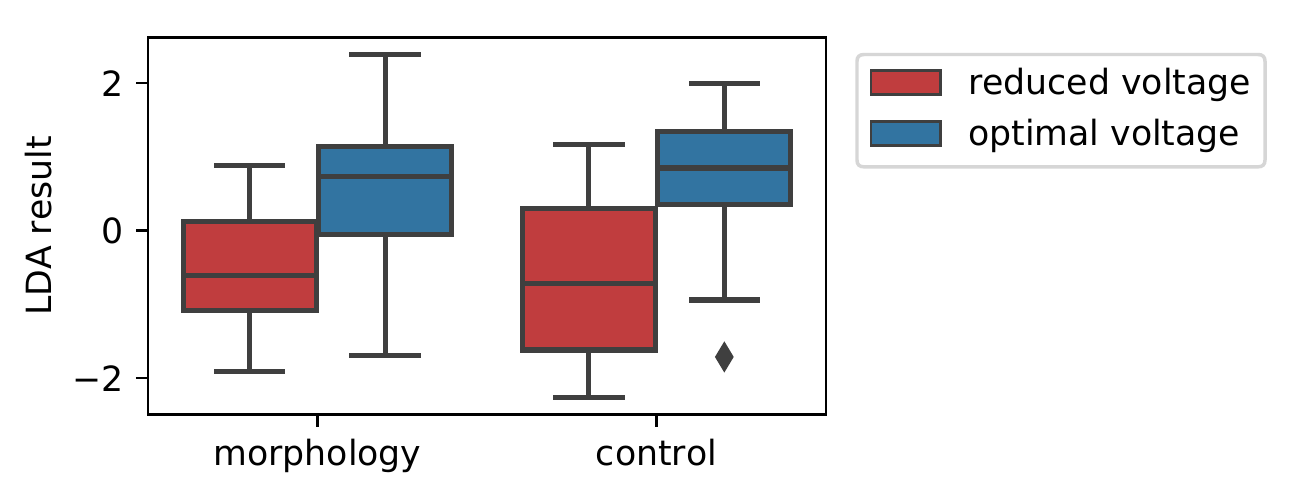}
    \caption{The resulting individuals when evolving under different internal environments \cite{tonnesfn_GECCO18}. We saw a significant difference in both morphology and control, showing that the search exploited both for adapting to the new environment.}
    \label{fig.expresults2}
  \label{fig.expresults}
  \vspace{-0.61cm}
\end{figure}

\textbf{Real-world evolution.}
We have also done evolutionary experiments on this platform, and again reduced the torque of the servos to see if the evolutionary process was able to adapt to this change~\cite{tonnesfn_GECCO18}.
We saw through our experiments that the evolutionary search exploited both control and morphology to adapt to the change in the internal environment.
We reduced the dimensionality of all morphology and control parameters respectively into two new dimensions using linear discriminant analysis, and saw a significant difference between the two environments for both dimensions, see Fig.\ref{fig.expresults2}.
Not only did we show that the system was able to adapt, we also demonstrated the feasibility of doing evolutionary experiments on the platform exclusively in the real world, completely bypassing reality gap effects.

\textbf{Simulation experiments.}
We have also developed a simulation model for the robot, which uses the Gazebo simulator and integrates with the ROS environment used for the other control functions of the robotic platform.
This is used in evolutionary experiments in~\cite{Nordmoen18}, where we focus on generating repertoires of low-level controller primitives, disconnected from the high-level control used in the hardware experiments.
While these results cannot currently be compared to the results from the hardware experiments, we emphasize the value of being able to leverage the performance of simulation to accompany experiments on a real-world platform.

\textbf{Discussion.}
By having a self-reconfigurable system where transitioning between morphologies is performed in a few seconds, and applying it to evolutionary robotics, we have shown that it is feasible to do experiments on a physically reconfigurable system.
Our typical evolutionary experiments with eight generations of eight individuals take about 90 minutes, including dynamic reconfiguration, servo cooldown and manual interventions for falls or other reasons.
This kind of extensive testing is not possible with other systems that feature manual or slow external reconfiguration.
By not using simulation, we bypass issues with the reality gap, but on the other hand, this may have exposed us even more to the issues of the complex search space of dynamic morphology optimization.
We consider our efforts to be a contribution towards the feasibility of real-world experiments on dynamic robot morphologies.
However, our robotic platform comes with a limited resolution of reconfiguration, i.e., it would be interesting to perform this kind of experiments also for a larger range of morphological configurations.

\subsection{Future experiments with our platform}

\textbf{Reality gap investigation.}
Being able to do hardware experiments on different morphologies on advanced four-legged robots opens up new avenues of investigating the reality gap and how it is affected by different bodies.
Not only is it useful to look into how the reality gap itself is affected by different morphologies, but also if a dynamic morphology can aid in some way to lessen the reality gap or its impact on the search process.

\textbf{Embodied cognition.}
The reality gap looks at the difference in performance when evaluating in the simulator versus the real world, but the concept of embodied cognition states that there is a bigger difference that might also make real world experiments more beneficial.
It states that the control, morphology, environment and the interactions between all these can serve as a source of cognition, and could be exploited to solve a task better.
Being able to test this in the rich and naturally noisy environment of the real world might yet yield a difference beyond the reality gap, showing the advantage of an embodied optimization  process.

\textbf{Control and morphology relationship.}
We have shown through earlier experiments that an evolutionary optimization of both control and morphology for different internal environments ended up significantly changing both, but a more thorough investigation is needed.
Comparing optimization or evolution of control, morphology, and both, might yield interesting insights into how a system is able to exploit the different aspects of the robot.
It would also be interesting to see how this would affect the search or optimization process, as the fitness landscape changes significantly in both size and complexity between these different approaches.

\textbf{Meta studies.}
Being able to test different control and morphology pairs efficiently in the real world opens up more realistic investigations into evolutionary processes in the field of artificial life.
Simulations are mostly used today, but lack the inherent noise and richness of the environment that only real-world experiments replicates.
It would limit the morphological complexity of the virtual creatures tested when compared to simulations, but might still lead to new and interesting results not seen in physics simulations alone.

\textbf{Automatic testing.}
Many experiments today are done on one robot, but the findings are being generalized to similar robots or even robotics in general.
Being able to use a robot like ours will allow testing or verification to be done with different leg lengths, so the jump to other similar robots become smaller.
It will also help in sharing research, as more results will not be overfitted to the single robot the researchers used, but to a range of morphologies available to them that is more likely to be usable by other robots.
Experiences from the use of this robot will hopefully also inspire more self-reconfigurable robot systems with other configurations or architectures.
Doing experiments on a range of different robot morphologies of different configurations will help researchers develop high-level algorithms and strategies that are more general and applicable to a wider audience in the real world.


\section{Conclusion and future work}

We have given an introduction to some of the challenges and possibilities from having a robot with dynamic morphology, and have introduced our take on a platform with self-reconfigurable morphology.
We introduced some of the experiments we have done, and what other experiments can be done with a simple platform like this.

Even though our platform gives new and exciting new ways to investigate and answer some of the problems and challenges in the field of dynamic morphology, it is just a start, and will serve as a stepping stone to new and better platforms in the future.
One of the main challenges with the level of technical complexity of current robots, are that they do not allow a very fine-grained resolution of the self-modifying hardware.
Several fields are working to get robots with more fine-grained and complex self-reconfigurable hardware, including modular robots, dynamic materials or soft robots.
We expect to see dynamic morphology being introduced to more sub-fields of robotics in the coming years as the technology matures, which will bring with it important perspectives and experience from other research fields.


\bibliographystyle{IEEEtran}
\bibliography{bibliography} 

\end{document}